\documentclass[11pt]{article}
\usepackage[margin=1in]{geometry}
\usepackage{amsmath,amssymb}
\usepackage{booktabs}
\usepackage{array}
\usepackage{tabularx}
\usepackage{float}
\usepackage{graphicx}
\usepackage{xcolor}
\usepackage{tikz}
\usepackage{hyperref}
\usepackage[authoryear,round]{natbib}
\usetikzlibrary{arrows.meta,positioning,fit,backgrounds,calc}

\hypersetup{colorlinks=true, linkcolor=blue, citecolor=blue, urlcolor=blue}
\floatstyle{ruled}
\newfloat{algorithm}{tbp}{loa}
\floatname{algorithm}{Algorithm}
\newcommand{\assetimage}[2]{\makebox[\linewidth][c]{\IfFileExists{#2}{\pdfximage width #1 {#2}\pdfrefximage\pdflastximage}{\fbox{\parbox{0.9\linewidth}{\centering Figure asset placeholder}}}}}
\newcommand{\best}[1]{\textbf{#1}}
\definecolor{elvblue}{HTML}{DCEEFF}
\definecolor{elvblueedge}{HTML}{4472A6}
\definecolor{elvgreen}{HTML}{E6F3DC}
\definecolor{elvgreenedge}{HTML}{5B8E4D}
\definecolor{elvorange}{HTML}{FFF0D8}
\definecolor{elvorangeedge}{HTML}{B5742E}
\definecolor{elvred}{HTML}{F8E0DE}
\definecolor{elvrededge}{HTML}{A9534B}
\definecolor{elvgray}{HTML}{F3F3F0}
\definecolor{elvlav}{HTML}{EEE7F6}
\definecolor{elvyellow}{HTML}{FFE9A8}
\definecolor{elvline}{HTML}{6D6D6D}
\definecolor{elvtext}{HTML}{222222}
\tikzset{
  elvbox/.style={draw=#1!85!black, fill=#1, rounded corners=3pt, line width=0.5pt, align=center, inner xsep=5pt, inner ysep=4pt, font=\rmfamily\itshape\scriptsize, text=elvtext},
  elvbox/.default=elvgray,
  elvsmall/.style={draw=#1!80!black, fill=#1, rounded corners=2pt, line width=0.42pt, align=center, inner xsep=4pt, inner ysep=3pt, font=\rmfamily\itshape\tiny, text=elvtext},
  elvsmall/.default=elvgray,
  elvarrow/.style={-{Latex[length=2.2mm]}, line width=0.55pt, draw=black!62},
  elvdash/.style={-{Latex[length=2.0mm]}, line width=0.52pt, draw=black!45, dashed},
  elvtitle/.style={font=\rmfamily\bfseries\itshape\tiny, text=black!82, align=center},
  elvlabel/.style={font=\rmfamily\itshape\tiny, text=black!72},
  eawmFrame/.style={draw=black!75, rounded corners=4pt, line width=0.75pt, fill=white},
  eawmPanel/.style={draw=#1!72!black, fill=#1!30, rounded corners=5pt, line width=0.65pt, inner sep=5pt},
  eawmBox/.style={draw=#1!78!black, fill=#1, rounded corners=3pt, line width=0.55pt, align=center, inner xsep=5pt, inner ysep=4pt, font=\rmfamily\bfseries\scriptsize, text=elvtext},
  eawmSmall/.style={draw=#1!75!black, fill=#1, rounded corners=2pt, line width=0.45pt, align=center, inner xsep=3pt, inner ysep=2.2pt, font=\rmfamily\scriptsize, text=elvtext},
  eawmTiny/.style={draw=#1!70!black, fill=#1, rounded corners=2pt, line width=0.4pt, align=center, inner xsep=2.5pt, inner ysep=1.8pt, font=\rmfamily\tiny, text=elvtext},
  eawmArrow/.style={-{Latex[length=2.3mm]}, line width=0.62pt, draw=black!68},
  eawmDash/.style={-{Latex[length=2.0mm]}, line width=0.55pt, draw=black!55, dashed},
  eawmBrace/.style={line width=0.55pt, draw=black!65},
}

\makeatletter
\def\ps@preprint{%
  \def\@oddhead{\small EV-WM\hfil A Preprint}%
  \def\@evenhead{\small EV-WM\hfil A Preprint}%
  \def\@oddfoot{\hfil\thepage\hfil}%
  \def\@evenfoot{\hfil\thepage\hfil}%
}
\let\ps@plain\ps@preprint
\makeatother
\makeatletter
\renewcommand{\@maketitle}{%
  \newpage
  \null
  \vspace*{-0.35in}%
  \begin{center}%
    \fontfamily{ptm}\selectfont
    \rule{0.88\textwidth}{1.1pt}\par
    \vspace{1.1em}%
    {\LARGE\bfseries\centering
      \begin{minipage}{0.92\textwidth}
      \centering\@title
      \end{minipage}\par}%
    \vspace{0.95em}%
    \rule{0.88\textwidth}{1.1pt}\par
    \vspace{0.85em}%
    {\large\scshape \@date\par}%
    \vspace{1.6em}%
    {\normalsize \@author\par}%
  \end{center}%
  \vspace{1.0em}%
}
\makeatother

\title{\textit{EV-WM}: Event-Verified World Models for Long-Horizon Robotic Manipulation}
\author{
\textbf{Kailin Wang\textsuperscript{1}} \quad
\textbf{Haoxiang Jie\textsuperscript{1}} \quad
\textbf{Yaoyuan Yan\textsuperscript{1}} \quad
\textbf{Jiacheng Zhou\textsuperscript{2,3}} \quad
\textbf{Zhiyou Heng\textsuperscript{1}}\\[0.35em]
{\small \textsuperscript{1}AI Lab, Country Garden Services Group \quad
\textsuperscript{2}Fudan University \quad
\textsuperscript{3}Omni AI}
}
\date{A Preprint}

\begin{document}
\maketitle

\begin{abstract}
Pretrained-feature world models provide a useful substrate for robot imagination, but visual or latent prediction alone does not determine whether an imagined future satisfies task-relevant predicates. Long-horizon manipulation requires progress signals that are relational, predicate-level, and physically grounded: whether an object has moved, whether a drawer or contact state has changed, whether a placement predicate is satisfied, and whether a candidate future is reliable enough for execution. We introduce \textbf{EV-WM}, a predicate-grounded verification framework for world-model planning. EV-WM rolls out candidate futures in pretrained visual-feature space, decodes them into structured event states, and scores them using task-progress, semantic-consistency, physical-feasibility, and uncertainty terms. The verifier guides sampling-based planning, gates candidate actions, and, in the contact-sensitive LIBERO wine-rack setting, selects among PPO-generated proposals. Across navigation, deformable-object, wall-constrained, and language-described manipulation studies, EV-WM shows that predicate-grounded verification can make feature-space world-model planning more interpretable and better aligned with task progress.
\end{abstract}

\section{Introduction}

World models are a central component of embodied intelligence. Early latent world models showed that compact imagined dynamics can support control from high-dimensional observations \citep{ha2018worldmodels}. Latent planning methods then demonstrated that learned dynamics can improve sample efficiency for decision making from pixels \citep{hafner2019planet}. Dreamer-style agents extended this idea by learning behaviors through latent imagination rather than using the model only for test-time planning \citep{hafner2020dreamer,hafner2023dreamerv3}. More recent interactive video models suggest that future prediction can scale beyond low-dimensional simulators toward richer interactive environments \citep{bruce2024genie,wu2024ivideogpt}. However, the representation optimized by many world models is not always the representation required for robot planning. A long-horizon manipulation policy needs more than a future image or dense feature vector. It must determine whether a drawer is open, whether a target object has moved, whether an object is on the required support, whether a proposed action violates task preconditions, and whether the imagined future advances the task.

The central limitation is a \emph{process mismatch} between visual-only world models and the structure of manipulation skill acquisition. Many robot world models follow a visual prediction loop: they take images or video features as the dominant input, predict future pixels or latent visual states, and choose actions according to those predictions. Human skill learning is more explicitly multimodal. An operator learning a new procedure combines demonstrations with written instructions, rules, ordered steps, safety constraints, and feedback from the evolving workspace. Consequently, predicting the appearance of a future scene is necessary but insufficient. Successful manipulation also requires reasoning about why an action is appropriate, which preconditions hold, what event should occur next, and whether the intended spatial relation or physical contact has been achieved.

Introducing events does not fully resolve this limitation if the events are inferred only from visual change. A purely visual event model can still ignore task rules, action history, and robot-state constraints. We therefore formulate event understanding as a task-specification grounding problem. Visual imagination is grounded jointly in benchmark task definitions, action history, and robot state, and it is decoded into explicit event states that can be verified during long-horizon planning. In LIBERO, the benchmark provides natural-language task descriptions, but our implementation instantiates the task specification through task identifiers, BDDL rules, and simulator-derived predicates rather than through a learned language encoder.

We propose \textbf{EV-WM: Event-Verified World Models}. EV-WM treats a visual-feature world model as a base imagination engine and adds a task-grounded event layer above it. For each candidate action sequence, the base model predicts future visual features. An event predictor maps the imagined future to task-relevant events, including object-state changes, spatial relations, affordance-like progress signals, and task-success predicates. A verifier then evaluates whether the predicted event state indicates task progress, semantic consistency, physical plausibility, and sufficient certainty. Planning uses a combined objective that retains the base visual-feature cost while rewarding verified event progress.

Our contributions are as follows:
\begin{itemize}
\item We formalize a predicate-grounded verification layer for pretrained-feature world models, separating latent future prediction from task-progress verification.
\item We instantiate EV-WM in a lightweight DINO-WM-style pipeline with simulator-derived event labels, supervised event prediction, ranked verifier scoring, and verifier-guided CEM planning.
\item We show that calibrated verifier-guided planning improves PointMaze random-target success from 0.90 to 0.94 while preserving dataset-goal performance.
\item We report Deformable and Wall-Single planning results, with retrieval-initialized EV-WM-CEM reaching 94\% success on Deformable e10 blocks and archive-validated EV-WM-CEM reaching 95\% success on Wall-Single.
\item We transfer the pipeline to LIBERO-goal, where check-success-aligned verification reaches AUC 0.993947 and a wine-rack-specific PPO proposal study improves H=20 online hybrid success to 97/100.
\end{itemize}

\section{Related Work}

\subsection{Latent and Model-Based World Models}

Early world-model systems learned compact latent dynamics that could be used by a controller instead of acting directly in pixel space \citep{ha2018worldmodels}. PlaNet used latent dynamics for planning from image observations, showing that model-based control can be practical in learned latent space \citep{hafner2019planet}. Dreamer learned policies through imagined latent rollouts, shifting model use from test-time planning toward policy optimization \citep{hafner2020dreamer}. DreamerV2 and DreamerV3 improved the robustness and domain coverage of this latent-imagination paradigm \citep{hafner2021dreamerv2,hafner2023dreamerv3}. DayDreamer brought Dreamer-style world models closer to physical robot learning \citep{wu2022daydreamer}. TransDreamer explored transformer dynamics for reinforcement learning with world models \citep{chen2022transdreamer}. EV-WM shares the goal of using learned dynamics for decision support, but changes what the planner reads from the model: in addition to predicted latent features, it uses verified event-level task progress.

\subsection{Predictive Representations and Pretrained Visual Features}

DINOv2 provides transferable frozen visual features for recognition and geometry-sensitive tasks \citep{oquab2023dinov2}. I-JEPA frames prediction as matching abstract feature representations rather than reconstructing pixels \citep{assran2023ijepa}. V-JEPA extends feature prediction to video, reinforcing the idea that useful predictive structure can reside in representation space \citep{bardes2024vjepa}. V-JEPA 2 and LeWorldModel further connect joint-embedding prediction with planning-oriented world models \citep{assran2025vjepa2,maes2026leworldmodel}. Genie and iVideoGPT show that interactive video prediction can support environment-like rollouts at scale \citep{bruce2024genie,wu2024ivideogpt}. RoboDreamer and Dream to Manipulate use imagined futures for robot manipulation and compositional generalization \citep{zhou2024robodreamer,barcellona2024dreamtomanipulate}. DreamGen, WorldEval, and Cosmos broaden the role of world models toward data generation, policy evaluation, and physical-AI foundation modeling \citep{jang2025dreamgen,li2025worldeval,nvidia2025cosmos}. EV-WM builds on this line of work by treating pretrained features as a substrate from which task-grounded events can be decoded and verified.

\subsection{Vision-Language-Action Policies}

SayCan grounds language instructions in affordance scores, showing how symbolic language goals can be filtered by action feasibility \citep{ahn2022saycan}. RT-1 demonstrates that transformer policies can scale language-conditioned robot control across many tasks \citep{brohan2022rt1}. Diffusion Policy represents visuomotor control as action generation through denoising, which is complementary to verifier-based action selection \citep{chi2023diffusionpolicy}. OpenVLA and \(\pi_0\) move vision-language-action policies toward open-source and generalist robot control \citep{kim2024openvla,black2024pi0}. FAST focuses on efficient action tokenization for VLA policies \citep{pertsch2025fast}. 3D-VLA, Mobile ALOHA, and Hi Robot extend VLA-style control toward 3D grounding, mobile manipulation, and hierarchical instruction following \citep{zhen2024threedvla,fu2024mobilealoha,shi2025hirobot}. EV-WM is complementary to these policies: it can serve as a lookahead module that proposes, scores, or gates candidate action windows before execution.

\subsection{Benchmarks, Sampling, and Policy Optimization}

LIBERO provides language-conditioned manipulation tasks with simulator state and success predicates, making it a useful benchmark for event supervision and verifier alignment \citep{liu2024libero}. We also evaluate PointMaze, Deformable, and Wall-Single to cover navigation, deformable-object manipulation, and wall-constrained control. The Cross-Entropy Method provides a sampling-based optimizer for candidate trajectory search \citep{rubinstein1999cem}. PPO provides a clipped policy-gradient objective that can improve an action proposal distribution from online rewards without unconstrained policy updates \citep{schulman2017ppo}. We use CEM as the controlled optimizer for the non-LIBERO studies and the LIBERO Goal10 baseline. PPO is introduced only for the LIBERO wine-rack task, where contact-sensitive placement benefits from a stronger proposal distribution.

\section{Problem Formulation}

We consider a robot receiving an observation history \(o_{t-k:t}\), proprioception \(p_{t-k:t}\), a task specification \(c\), and candidate actions \(a_{t:t+H-1}\). A pretrained visual encoder \(E\) maps observations to features \(z_t = E(o_t)\). A base action-conditioned world model \(F_\theta\) predicts future visual features:
\[
\hat{z}_{t+H} = F_\theta(z_{t-k:t}, p_{t-k:t}, a_{t:t+H-1}).
\]
Feature-only planning selects actions by minimizing a latent cost, such as distance to a goal feature \(z_g\). EV-WM instead introduces an event state
\[
e_t = \{b_t, r_t, q_t, s_t, u_t\},
\]
where \(b_t\) denotes binary object and relation events, \(r_t\) denotes continuous distances or margins, \(q_t\) denotes joint or contact progress, \(s_t\) denotes task success or subgoal completion, and \(u_t\) denotes uncertainty. An event predictor \(G_\psi\) maps imagined features and task context to a future event state:
\[
\hat{e}_{t+H} = G_\psi(\hat{z}_{t+H}, z_{t-k:t}, c, a_{t:t+H-1}).
\]
The verifier scores the candidate future:
\[
S_{\mathrm{EV\text{-}WM}} = S_{\mathrm{task}} + \lambda_s S_{\mathrm{semantic}} + \lambda_p S_{\mathrm{physical}} - \lambda_u U.
\]
Planning minimizes
\[
J(a_{t:t+H-1}) = w_f C_{\mathrm{feature}} - S_{\mathrm{EV\text{-}WM}},
\]
where \(w_f\) calibrates the feature cost against the verifier score. This objective preserves the base visual rollout objective while making action selection sensitive to task-event progress.

\section{Method: Event-Verified World Models}

EV-WM consists of four components: a pretrained-feature world model, an automatic event-labeling interface, a task-grounded event predictor and verifier, and a planner that uses verified event progress for action selection. The following subsections describe event supervision, event prediction and verification, verifier-guided CEM, and the wine-rack PPO proposal policy used for contact-sensitive LIBERO placement.

\subsection{Architecture Overview}

\begin{figure}[!htbp]
\centering
\includegraphics[width=\linewidth]{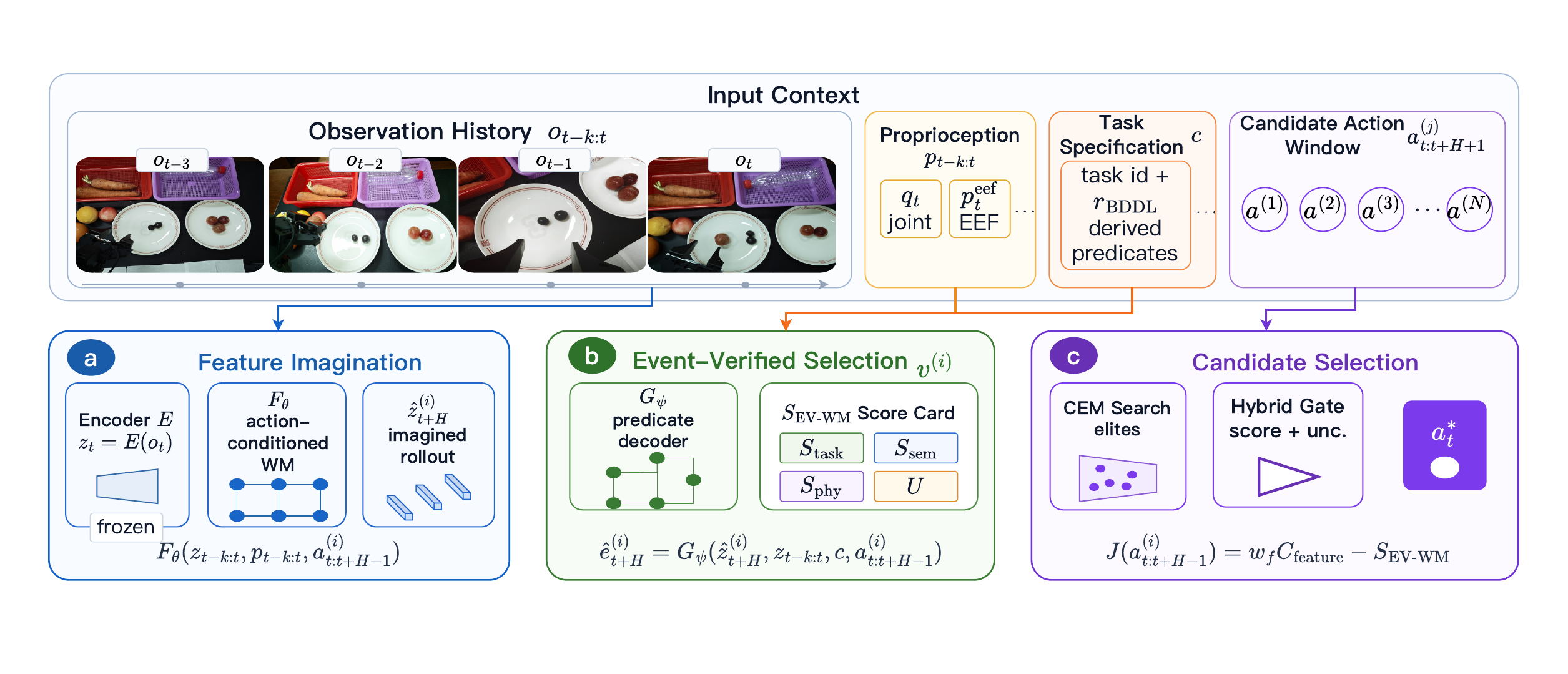}
\caption{Entity-grounded EV-WM overview. The input context combines recent observations \(o_{t-k:t}\), proprioception \(p_{t-k:t}\), task predicates from \(c\), and candidate action windows \(a_{t:t+H-1}^{(i)}\). Panel (a) imagines latent futures with a frozen visual encoder and an action-conditioned feature world model. Panel (b) decodes each imagined future into task-grounded event predicates and a verifier score card \(S_{\mathrm{EV\text{-}WM}}\). Panel (c) uses CEM elites and a conservative hybrid gate to select an executable action \(a_t^\ast\), so actions are selected by predicted event consistency rather than feature similarity alone.}
\label{fig:entity_grounded_prove}
\end{figure}

A frozen visual encoder extracts observation features, and an action-conditioned dynamics model rolls out future latent states under candidate actions. The base model is independent of the event verifier and can be reused from a DINO-WM-style feature-prediction pipeline. EV-WM does not require the visual model to reconstruct pixels or expose symbolic state. It only requires imagined latent futures to retain enough task information for a task-grounded event predictor to decode.

For the DINO-WM baselines, we use the same frozen encoder, action-conditioned rollout model, action horizon, CEM sampling budget, and action parameterization as the corresponding EV-WM setting. The baseline objective removes the event predictor, ranked verifier, and conservative gate, and optimizes only the visual-feature or state-distance cost \(C_{\mathrm{feature}}\), such as the distance between the predicted terminal representation and the dataset goal feature or target state. Thus, matched DINO-WM and EV-WM rows differ in the predicate-grounded scoring layer rather than in the rollout budget.

\subsection{Event Label Generation}

A key component of EV-WM is that event supervision is generated from simulator state and task definitions rather than manually annotated frame labels. Given a task-specific simulator state \(s_t\), initial state \(s_0\), and parsed BDDL rule set \(r_{\mathrm{BDDL}}\), we define the event labeler as a deterministic map
\[
\ell_t = g_{\mathrm{task}}(s_0, s_t, r_{\mathrm{BDDL}}),
\]
where \(\ell_t\) contains object-state events, relation events, continuous progress margins, and task-level success. In PointMaze, where the state is an agent coordinate \(x_t\), the labels are generated by
\[
d_{t+H}^{g} = \left\|x_{t+H} - x_g\right\|_2,\qquad
\rho_{t+H} =
\frac{\left\|x_t - x_g\right\|_2 - \left\|x_{t+H} - x_g\right\|_2}
{\left\|x_t - x_g\right\|_2 + \epsilon},
\]
\[
\ell_{t+H}^{\mathrm{success}} = \mathbb{I}\!\left[d_{t+H}^{g} < \tau_g\right],
\]
where \(x_g\) is the goal coordinate, \(\rho_{t+H}\) is normalized progress, and \(\tau_g=0.5\). In LIBERO-goal, the same principle is applied to object and joint states recovered by simulator replay:
\[
\ell_t^{\mathrm{moved}} =
\mathbb{I}\!\left[\left\|p_{\mathrm{obj}}(t)-p_{\mathrm{obj}}(0)\right\|_2 > \tau_{\mathrm{move}}\right],
\qquad
\ell_t^{\mathrm{near}} =
\mathbb{I}\!\left[\left\|p_{\mathrm{obj}}(t)-p_{\mathrm{eef}}(t)\right\|_2 < \tau_{\mathrm{eef}}\right],
\]
\[
\ell_t^{\mathrm{on}} =
\mathbb{I}\!\left[\left\|p_{\mathrm{obj}}(t)-p_{\mathrm{target}}(t)\right\|_2 < \tau_{\mathrm{place}}\right],
\qquad
\ell_t^{\mathrm{joint}} =
\mathbb{I}\!\left[\sigma_j q_j(t) > \tau_j\right].
\]
Here \(p_{\mathrm{obj}}\), \(p_{\mathrm{target}}\), and \(p_{\mathrm{eef}}\) denote object, target-site, and end-effector positions; \(q_j\) is a task-relevant drawer or stove joint; and \(\sigma_j\) encodes the task-specific opening or activation direction. Composite events are formed using task logic, e.g.,
\[
\ell_t^{\mathrm{composite}} =
\ell_t^{\mathrm{joint}} \wedge \ell_t^{\mathrm{on}},
\]
while the final task success label used by the verifier and online evaluation is aligned to LIBERO's native predicate,
\[
\ell_t^{\mathrm{success}} = \mathrm{check\_success}(s_t).
\]
This alignment keeps the verifier target consistent with the online evaluation criterion.

\subsection{Event Prediction and Verification}

The supervised event predictor \(G_\psi\) is trained from rollout-window samples \((z_{t-k:t}, c, a_{t:t+H-1}, \ell_{t+H})\), where \(\ell_{t+H}\) is generated by the simulator-derived labeler described above. We use binary cross-entropy for binary event predicates, squared or smooth-\(L_1\) regression for continuous distances and margins, and binary classification for task success. Positive and negative targets are obtained automatically from \texttt{check\_success}, predicate satisfaction, and progress margins; no manual frame-level annotation is required. In PointMaze, \(G_\psi\) outputs future position, distance to goal, progress to goal, success probability, and region class. In LIBERO-goal, it outputs object movement, object-near-end-effector, object lifted, object-on-target, drawer open, stove on, subgoal completion, object movement distance, task margins, and contact-related indicators.

For LIBERO-goal, we also train a ranked verifier/scorer on paired windows. Demonstration or successful windows are ranked above Gaussian, zero-action, shuffled, or lower-score candidate windows using a pairwise ranking loss. This ranked objective complements the supervised predicate losses by learning the relative ordering used by CEM and the conservative hybrid gate. The uncertainty term \(U\) is implemented as a low-confidence penalty rather than a strictly Bayesian uncertainty estimate. In our experiments, it can be approximated by success-probability entropy, margin-to-threshold confidence, or ensemble/dropout variance, and is used to reject low-confidence candidates in the conservative gate.

The verifier is not a second dynamics model; it scores the event predictor's output. In PointMaze, the task score combines success probability, progress, and predicted goal distance:
\[
S_{\mathrm{task}} = 0.45s + 0.35p + 0.20\exp(-d/d_{\max}).
\]
The total verifier score combines task, semantic, physical, and uncertainty terms:
\[
S_{\mathrm{total}} = 1.0S_{\mathrm{task}} + 0.5S_{\mathrm{semantic}} + 0.25S_{\mathrm{physical}} - 0.1U.
\]
Here \(S_{\mathrm{task}}\) measures task completion evidence, such as success probability, progress, goal distance, or predicate completion. \(S_{\mathrm{semantic}}\) measures task-logic consistency, including BDDL predicate compatibility, subgoal ordering, and required relation consistency. \(S_{\mathrm{physical}}\) measures physical feasibility through distance and contact margins, joint-direction consistency, and contact-sensitive placement checks. \(U\) penalizes uncertain or low-confidence event predictions. In PointMaze, the semantic and physical terms reduce to goal-region consistency and distance/progress consistency. In LIBERO-goal, they are instantiated from BDDL predicates, native \texttt{check\_success} alignment, object-relation constraints, contact margins, and task-relevant joint constraints. These are verifier scoring terms applied to predicted events; they do not introduce an additional dynamics model. At test time, CEM samples candidate action windows, rolls them out in feature space, predicts events, and scores the candidates. The selected CEM action is either executed directly or passed through a conservative hybrid gate that accepts CEM only when its score margin over the demonstration action is sufficiently reliable.

\subsection{Verifier-Guided CEM Planning}

For PointMaze, Deformable, Wall-Single, and the first LIBERO-goal online evaluation, EV-WM uses CEM as a controlled sampling-based planner. At each decision point, CEM samples action windows, rolls them out through the feature world model, predicts events for the imagined terminal state, and ranks candidates by the combined objective
\[
J(a_{t:t+H-1}) = w_f C_{\mathrm{feature}}(a_{t:t+H-1}) - S_{\mathrm{EV\text{-}WM}}(a_{t:t+H-1}).
\]
The elite candidates update the sampling distribution for the next CEM iteration. In online LIBERO evaluation, the selected CEM action can either be executed directly or passed through a hybrid gate that compares the CEM score with the demonstration-action score. This gate is intentionally conservative: it accepts CEM only when the verifier margin and task-specific safety conditions indicate that the imagined improvement is likely to transfer to simulator execution.

\subsection{Planner-Aware PPO Proposal}

The wine-rack task uses a residual PPO proposal policy to generate contact-sensitive candidates beyond the CEM pool. The policy samples an action-window residual around the demonstration action,
\[
a_{t:t+H-1}^{\mathrm{ppo}} = a_{t:t+H-1}^{\mathrm{demo}} + \Delta_\phi(o_{t-k:t},p_{t-k:t},g),
\]
where \(\Delta_\phi\) is clipped to a bounded residual range. Each sampled window is executed in the LIBERO simulator and receives a terminal reward
\[
r = \mathbb{I}\!\left[\mathrm{env.check\_success}()=\mathrm{true}\right].
\]
The proposal policy uses the PPO clipped surrogate objective:
\[
L^{\mathrm{clip}}(\phi)=
\mathbb{E}_t\!\left[
\min\left(
\rho_t(\phi)\hat{A}_t,\;
\mathrm{clip}(\rho_t(\phi),1-\epsilon,1+\epsilon)\hat{A}_t
\right)
\right],
\]
where \(\rho_t(\phi)=\pi_\phi(\Delta_t\mid o,p,g)/\pi_{\phi_{\mathrm{old}}}(\Delta_t\mid o,p,g)\), \(\hat{A}_t\) is the advantage estimate, and \(\epsilon\) is the PPO clipping parameter. In practice, this PPO policy is not used as an unconstrained replacement for the demonstration action. It produces a candidate set around the demonstration window, and the verifier/reranker acts as a deployable selector over the top-ranked PPO candidates rather than as an oracle.

\begin{algorithm}[H]
\centering
\footnotesize
\setlength{\tabcolsep}{3pt}
\caption{Planner-aware PPO proposal with predicate-grounded verifier selection. Residual action proposals are scored through a verifier/reranker gate rather than directly executing the policy top-1 action.}
\label{alg:ppo_proposal}
\begin{tabularx}{0.96\linewidth}{>{\raggedright\arraybackslash}p{0.16\linewidth}>{\raggedright\arraybackslash}X}
\toprule
\textbf{Stage} & \textbf{Pseudocode} \\
\midrule
\textbf{Input} & Demonstration action window \(a^{\mathrm{demo}}\), residual policy \(\pi_\phi\), verifier score \(S_{\mathrm{EV\text{-}WM}}\), candidate reranker, and task predicate \(\mathrm{check\_success}\). \\
\textbf{Train} & For each demonstration window, sample residuals \(\Delta\), form \(a^{\mathrm{ppo}}=a^{\mathrm{demo}}+\Delta\), clip residual actions to the allowed range, execute candidates in the simulator, and assign terminal reward \(r=\mathbb{I}[\mathrm{check\_success}]\). \\
\textbf{Update} & Estimate advantages \(\hat{A}\) from terminal rewards and update \(\pi_\phi\) with the PPO clipped objective \(L^{\mathrm{clip}}\). \\
\textbf{Deploy} & Sample \(K\) PPO candidates, score them with the verifier/reranker, and accept a PPO candidate only when the top-2 gate passes; otherwise execute the demonstration window. \\
\bottomrule
\end{tabularx}
\end{algorithm}

This design combines proposal generation, event-verifier selection, and predicate-aligned execution in a deployable H=20 action-window policy.

For placement tasks, this execution detail is important. In the wine-rack task, the true goal is an \(\mathrm{On}(\mathrm{wine\_bottle},\mathrm{wine\_rack\_top\_region})\) predicate, which requires both target-region containment and rack-bottle contact. We therefore add a predicate-informed final settle tail only for this contact-sensitive evaluation:
\[
a_{t+i}^{\mathrm{settle}} = (\Delta z=-0.05,\; \mathrm{gripper}=-1),\qquad i=1,\ldots,N_{\mathrm{settle}}.
\]
This tail is treated as task-conditional execution alignment rather than a generic policy improvement. The final deployable selector applies the candidate reranker only within the top two PPO candidates, improving placement success while preserving the demonstration fallback.

\section{Experiments}

\subsection{Benchmarks}

We evaluate EV-WM on four complementary benchmarks. \textbf{PointMaze} is a continuous 2D navigation environment in which an agent must reach a target location from a given state. It provides a compact testbed for evaluating whether predicate-grounded verifier scores improve feature-space planning, because success can be measured directly by goal distance and random target states test whether the planner generalizes beyond dataset goals.

\textbf{Deformable} evaluates manipulation of a deformable object whose state changes are difficult to summarize with a single rigid-body pose. The task stresses long-horizon action selection and initialization quality: a planner must produce action sequences that guide the object toward the target shape or configuration, and success is measured by task completion together with geometric distance metrics such as Chamfer distance. We use this benchmark to test whether EV-WM can convert retrieved latent action priors into effective online planning behavior.

\textbf{Wall-Single} is a wall-constrained control benchmark in which the agent must reach target states while respecting the structure imposed by a single obstacle or wall. It is useful for evaluating candidate selection because visually plausible or feature-close rollouts can still be poor choices if they violate the constrained transition geometry. We use this benchmark to test whether predicate-grounded scoring selects better candidates than visual-feature cost alone.

\textbf{LIBERO-goal} is a language-described robotic manipulation benchmark with simulator state, BDDL task definitions, and native \texttt{check\_success} predicates. It covers object placement, articulated-object interaction, and contact-sensitive manipulation under natural-language task descriptions. In our experiments, these descriptions are used through benchmark task names and BDDL-derived predicates, not through a learned language encoder. We use Goal10 to evaluate event verification, CEM, and conservative hybrid gating across multiple task-specified manipulation problems, and we use the wine-rack task as a contact-sensitive setting for PPO proposal generation.

Across these benchmarks, \textbf{PointMaze} tests calibrated verifier-guided planning, \textbf{Deformable} tests retrieval-initialized deformable-object planning, \textbf{Wall-Single} tests archive-validated candidate selection, and \textbf{LIBERO-goal} tests task-grounded predicate verification for language-described manipulation tasks. PointMaze, Deformable, and Wall-Single use CEM, retrieval, and validation-style selection; they do not use PPO.

The LIBERO evaluation contains two settings. Goal10 compares demonstration replay, demo-initialized CEM, and conservative hybrid gating across task-specified manipulation problems. PPO is used only for the contact-sensitive \texttt{put\_the\_wine\_bottle\_on\_the\_rack} task, where stronger proposal generation improves H=20 placement performance.

\section{Results and Analysis}

\subsection{PointMaze Planning}

Calibrated event verification improves the random-target PointMaze setting while preserving performance on dataset goals. Table~\ref{tab:pointmaze_main} summarizes the main comparison, and Table~\ref{tab:feature_weight} shows that the verifier score must be balanced against the feature cost.

\begin{table}[H]
\centering
\small
\caption{PointMaze planning results. Rows are separated into matched comparison blocks with the same goal source and planning horizon. Bold marks the better value within each block: higher is better for success rate, and lower is better for mean state distance. Dataset-goal success is saturated, so random-state planning provides the more informative comparison.}
\label{tab:pointmaze_main}
\begin{tabular}{llcccc}
\toprule
Method & Goal source & Goal H & Feature weight & Success rate & Mean state distance \\
\midrule
DINO-WM & dataset & 5 & -- & 1.00 & 0.64862 \\
EV-WM & dataset & 5 & 1 & 1.00 & \best{0.63997} \\
\midrule
DINO-WM & dataset & 10 & -- & 1.00 & \best{0.59562} \\
EV-WM & dataset & 10 & 1 & 1.00 & 0.63262 \\
\midrule
DINO-WM & random state & 5 & -- & 0.90 & 0.93568 \\
EV-WM & random state & 5 & 10 & \best{0.94} & \best{0.90573} \\
\bottomrule
\end{tabular}
\end{table}

\begin{table}[H]
\centering
\caption{Feature-weight ablation on PointMaze random-state planning. The DINO-WM row is the no-verifier baseline, and the EV-WM rows vary the feature-cost weight. Bold marks the best value in each metric column, showing that success and final distance peak at different feature weights.}
\label{tab:feature_weight}
\begin{tabular}{lccc}
\toprule
Method & Feature weight & Success rate & Mean state distance \\
\midrule
DINO-WM & -- & 0.90 & 0.93568 \\
\midrule
EV-WM & 1 & 0.84 & 1.02408 \\
EV-WM & 5 & 0.92 & \best{0.89129} \\
EV-WM & 10 & \best{0.94} & 0.90573 \\
EV-WM & 20 & 0.88 & 1.01147 \\
\bottomrule
\end{tabular}
\end{table}

On dataset goals, the baseline already reaches 1.00 success, and EV-WM preserves this behavior. On random-state goals, EV-WM improves success from 0.90 to 0.94 and reduces mean state distance after calibration. The ablation indicates that an overly dominant verifier score can hurt planning when its scale overwhelms the feature cost. Moderate feature weighting allows event verification to improve action selection without dominating the rollout objective.

\subsection{Deformable Retrieval-Initialized Planning}

The Deformable setting evaluates predicate-grounded planning with a world-model latent retrieval prior. Retrieval provides a task-relevant action initialization, and conservative EV-WM-CEM performs local optimization around that trajectory. The comparison separates planning without an action prior, local optimization around successful demonstration actions, and online planning initialized from the nearest latent trajectory.

\begin{table}[H]
\centering
\small
\caption{Deformable planning results. Rows are separated by evaluation scope. The local pass-rate row is a sanity check and is not directly comparable to online task success; the online rows report deployable planning settings. Bold marks the primary online-planning result.}
\label{tab:deformable}
\begin{tabularx}{0.94\linewidth}{>{\centering\arraybackslash}m{0.21\linewidth}>{\centering\arraybackslash}m{0.25\linewidth}>{\centering\arraybackslash}m{0.17\linewidth}>{\centering\arraybackslash}X}
\toprule
Setting & Evaluation scope & Result & Interpretation \\
\midrule
\shortstack[c]{Zero-init\\EV-WM-CEM} & \shortstack[c]{online planning\\(online-comparable)} & \shortstack[c]{0\%\\success} & \shortstack[c]{No retrieval\\prior} \\
\midrule
\shortstack[c]{Demo-local CEM /\\EV-WM-CEM} & \shortstack[c]{local action validation\\(sanity check only)} & \shortstack[c]{100\%\\local pass rate} & \shortstack[c]{Preserves successful\\local actions} \\
\midrule
\shortstack[c]{Nearest-latent init +\\conservative\\EV-WM-CEM} & \shortstack[c]{e10 online planning\\(online-comparable)} & \shortstack[c]{\best{94\%}\\\best{success}} & \shortstack[c]{Primary deployable\\online setting} \\
\bottomrule
\end{tabularx}
\end{table}

The Deformable results show that predicate-grounded planning is most effective when paired with a task-relevant action prior. Local CEM / EV-WM-CEM preserves successful demonstration-neighborhood actions with a 100\% local pass rate, while retrieval-initialized conservative EV-WM-CEM reaches 94\% success on the e10 online evaluation. The contrast with zero initialization indicates that retrieved latent actions and verifier scoring jointly provide task-progress guidance beyond visual-feature prediction alone.

\subsection{Wall-Single Archive Validation}

Wall-Single provides a stress test for predicate-grounded scoring and candidate selection. The DINO-WM MPC-CEM baseline reaches 0.88 success on 50 random-state goals. EV-WM-CEM improves both the final policy and archive-based candidate selection, with the strongest configuration reaching 95\% success. This result suggests that verifier scoring can outperform the visual-feature baseline when paired with a reliable candidate-selection protocol.

\begin{table}[H]
\centering
\small
\caption{Wall-Single random-state planning. The first row is the DINO-WM MPC-CEM baseline; the following rows add EV-WM-CEM and archive validation. Bold marks the best online planning result, where higher success and lower distance are both better.}
\label{tab:wall}
\begin{tabularx}{0.96\linewidth}{>{\raggedright\arraybackslash}X c c c}
\toprule
Method & Evaluations & Success rate & Mean state distance \\
\midrule
DINO-WM MPC-CEM baseline & 50 & 0.88 & 3.75449 \\
\midrule
EV-WM-CEM fw5 final & 50 & 0.92 & 3.54676 \\
EV-WM-CEM fw5 early10 archive top50 validation & 50 & \best{0.95} & \best{3.23375} \\
\bottomrule
\end{tabularx}
\end{table}

The calibrated EV-WM-CEM final policy improves over the DINO-WM baseline, reaching 0.92 success and reducing mean state distance to 3.54676. Early archive top50 validation further raises success to 0.95 and lowers mean state distance to 3.23375. In this setting, EV-WM provides a stronger planning signal than visual-feature cost alone because it preserves useful early candidates and selects actions according to verified task progress.

\subsection{LIBERO-goal Results}

On imagined LIBERO-goal rollouts, native \texttt{check\_success} alignment yields a discriminative ranked verifier, and offline CEM improves the learned combined score over demonstration actions in most evaluated windows.

\begin{table}[H]
\centering
\small
\caption{LIBERO-goal offline verifier, planning-score, and CEM comparisons after check-success label alignment. Horizontal rules separate verifier metrics, planning-score sanity checks, and offline CEM tests. Bold marks the best value within each separated comparison block; the two CEM rows are both bold because they report the CEM improvement rate and average improvement margin.}
\label{tab:offline_verifier}
\begin{tabular}{lrlr}
\toprule
Module & Samples/windows & Metric & Result \\
\midrule
Supervised verifier & 1600 & AUC & 0.991625 \\
Ranked verifier & 1600 & AUC & \best{0.993947} \\
\midrule
Supervised verifier & 1600 & Score gap & \best{1.274701} \\
Ranked verifier & 1600 & Score gap & 1.236445 \\
\midrule
Planning score & 1600 & Demo \(>\) Gaussian & \best{0.998125} \\
Planning score & 1600 & Demo \(>\) Zero & 0.983750 \\
Planning score & 1600 & Demo \(>\) Shuffle & 0.866250 \\
\midrule
Offline CEM & 200 & CEM \(>\) Demo & \best{0.895000} \\
Offline CEM & 200 & CEM - Demo & \best{+0.056604} \\
\bottomrule
\end{tabular}
\end{table}

Native \texttt{check\_success} alignment gives the verifier the same success definition used by online LIBERO evaluation. This avoids overly permissive geometric labels and keeps the offline verifier, offline CEM score, and H=20 online evaluation aligned to the same predicate-level target.

Online evaluation executes H=20 action windows in the LIBERO simulator and evaluates the resulting state with \texttt{env.check\_success()}. This is a short-window evaluation rather than a full autonomous long-horizon episode rollout. We report the Goal10 CEM/hybrid comparison as the baseline online manipulation setting, and then use wine-rack placement as a contact-sensitive test for PPO proposal generation. The wine-rack task evaluates planner-aware PPO proposals for a bottle-on-rack placement task whose success is defined by LIBERO's true \texttt{On} predicate.

The wine-rack task benefits from contact-sensitive proposals and conservative verifier selection. Without a settle tail, the deployable hybrid reaches 70/100. With a 9-step predicate-informed settle tail, PPO top1 reaches 96/100, and the top-2 verifier/reranker hybrid reaches 97/100, matching Oracle\mbox{@}32 and improving over the 92/100 demonstration baseline. The Goal10 online comparison is more modest: conservative verification improves demonstration replay from 87/100 to 88/100 while avoiding direct CEM replacement, which drops to 75/100. Overall, EV-WM acts as a conservative predicate-grounded selection layer. It preserves reliable demonstration actions when generated candidates are weak and exploits stronger proposal distributions when predicate-level evidence supports them.

\begin{figure}[H]
\centering
\makebox[\linewidth][c]{\includegraphics[page=2,width=1.08\linewidth,trim=0 42pt 0 6pt,clip]{draw/libero_success_bar.drawio.pdf}}
\caption{Wine-rack H=20 online PPO proposal success rates. The top-2 verifier hybrid reaches 97/100, matching Oracle\mbox{@}32.}
\label{fig:libero_success_bar}
\vspace{0.35em}
\centering
\includegraphics[width=0.94\linewidth]{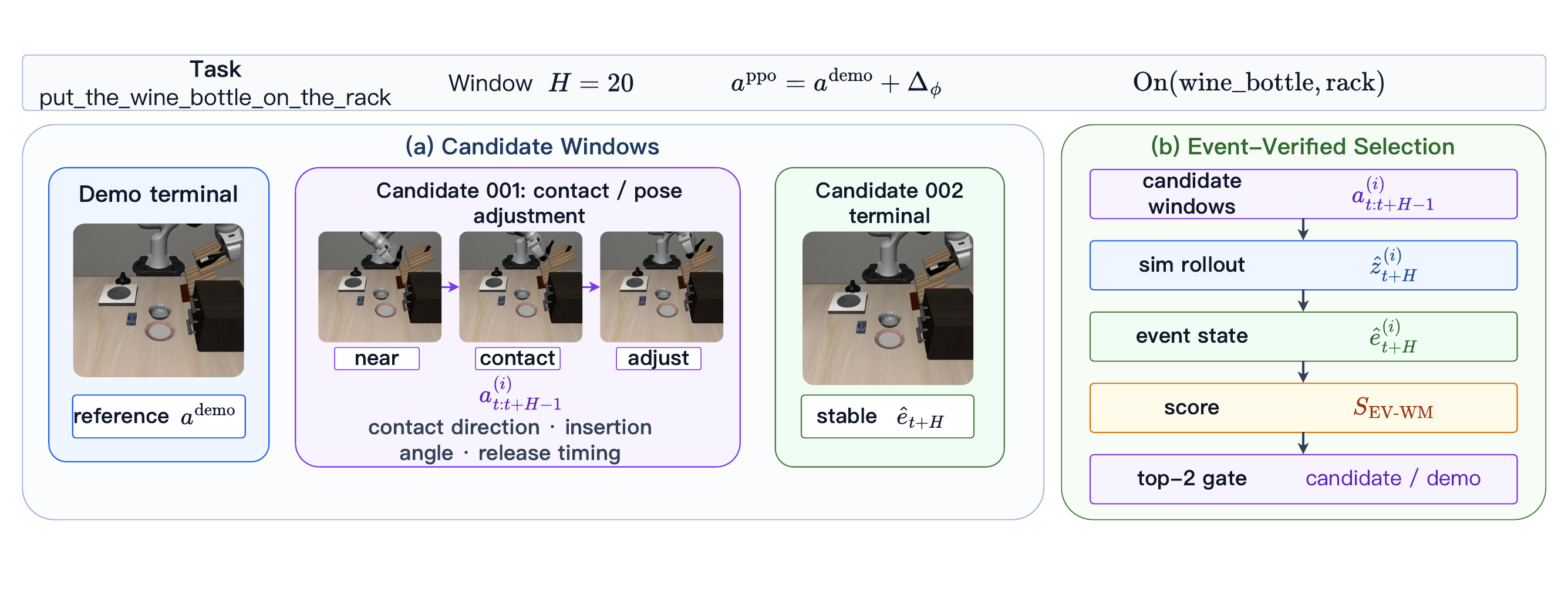}
\caption{LIBERO wine-rack simulation evidence and event-verifier selection. Candidate 001 exposes the contact and pose-adjustment process that makes the task sensitive to predicate-level verification.}
\label{fig:wine_candidate_selection}
\end{figure}

\section{Limitations}

EV-WM relies on simulator-derived event supervision. This is appropriate for a controlled first study, but real-world deployment would require robust perception, VLM-assisted labeling, manual auditing, or other forms of event extraction. The LIBERO online evaluation is also a short-window H=20 test rather than full episode-level autonomous execution. The offline verifier and CEM scorer improve imagined-rollout scores, while the online Goal10 results indicate that conservative hybrid gating is more reliable than direct CEM replacement. The PPO proposal study focuses on one wine-rack task, and its settle tail is a predicate-informed execution protocol for contact-sensitive placement rather than a general policy improvement. The Deformable and Wall-Single results are controlled evaluations under retrieval and archive-validation protocols, so broader deployable benchmark validation remains necessary. Finally, the verifier is partly rule-structured and partly learned; dense verifier learning, task-conditional execution alignment, target-region-aware proposal generation, and full long-horizon autonomous evaluation remain important directions for future work.

\section{Conclusion}

We presented EV-WM, a framework that augments pretrained visual-feature world models with predicate-grounded event prediction and verification. The core idea is to plan not only toward visually plausible futures, but also toward futures that make verified task-event progress. PointMaze experiments show that verifier-guided CEM improves random-target planning from 0.90 to 0.94 success when the feature and verifier objectives are calibrated. Deformable reaches 94\% success with retrieval-initialized conservative EV-WM-CEM, and Wall-Single reaches 95\% success with early archive top50 validation. LIBERO-goal experiments show that native-predicate event verification reaches AUC 0.993947 and that conservative hybrid online gating improves Goal10 H=20 execution from 87/100 to 88/100. The wine-rack PPO proposal study further indicates that predicate-grounded selection can support contact-sensitive placement, reaching 97/100 and matching Oracle\mbox{@}32. These results support EV-WM as an interpretable predicate-grounded verification layer for feature-space world models, with future work focused on dense verifier learning, target-region-aware proposal generation, task-conditional predicate handling, and full long-horizon autonomous evaluation.

\bibliographystyle{plainnat}
\begingroup
\raggedright
\bibliography{references}
\endgroup

\end{document}